\documentclass[sigconf]{acmart}
\usepackage{multirow}
\usepackage{bbding}
\usepackage[misc]{ifsym}
\usepackage{amsmath}
\usepackage{pifont}

\newcommand{\cmark}{\ding{51}}%
\AtBeginDocument{%
  }

\begin{document}
\renewcommand\footnotetextcopyrightpermission[1]{}
\settopmatter{printacmref=false}

\title{EgoAnimate: Generating Human Animations from Egocentric top-down Views}

\author{G. Kutay Türkoglu}

\orcid{0009-0009-9780-5721}
\affiliation{%
  \institution{Sony Semiconductor Solutions Europe}
  \country{Germany}
}
\email{Guerbuez.Tuerkoglu@sony.com}

\author{Julian Tanke}
\affiliation{%
  \institution{Sony AI}
  \country{\country{United States}}
  }
\email{Julian.Tanke@sony.com}

\author{Iheb Belgacem}
\affiliation{%
  \institution{Sony Semiconductor Solutions Europe}
  \country{Germany}
}
\email{Iheb.Belgacem@sony.com}

\author{Lev Markhasin}
\affiliation{%
  \institution{Sony Semiconductor Solutions Europe}
  \country{Germany}
}
\email{Lev.Markhasin@sony.com}
\settopmatter{
  printacmref=false,    
  printccs=false,        
}
\renewcommand\footnotetextcopyrightpermission[1]{} 


\begin{abstract}
An ideal digital telepresence experience requires the accurate replication of a person's body, clothing, and movements. In order to capture and transfer these movements into virtual reality, the egocentric (first-person) perspective can be adopted, which makes it feasible to rely on a portable and cost-effective stand-alone device that requires no additional front-view cameras. However, this perspective also introduces considerable challenges, particularly in learning tasks, as egocentric data often contains severe occlusions and distorted body proportions.
 
There are only a few works that perform human appearance reconstruction from egocentric videos, and none of them adopt a generative prior-based approach. Although some methods can generate an avatar from a single egocentric input at inference time, their training processes still rely on multi-view motion-capture datasets. To the best of our knowledge, this is the first study to utilize a generative backbone for the reconstruction of animatable avatars from egocentric inputs. Using a Stable Diffusion-based prior, our method reduces the training burden and improves generalizability.
 
The idea of synthesizing fully occluded parts of an object has been widely explored in various domains. In particular, models such as SiTH and MagicMan have demonstrated successful 360-degree reconstruction from a single frontal image. Inspired by these approaches, we propose a pipeline that reconstructs a realistic frontal view from a highly occluded top-down image using ControlNet and a Stable Diffusion backbone enabling the synthesis of novel views.
 
Our primary goal is to transform a single egocentric top-down image of a person into a realistic frontal representation and feed it into a state-of-the-art image-to-motion model. This enables the generation of desired avatar motions from limited egocentric inputs, paving the way for more accessible and generalizable telepresence systems.
\end{abstract}



\keywords{Egocentric Camera View, Human Avatar}
\captionsetup[figure]{labelfont=bf,textfont=normalfont}
\begin{teaserfigure}
  \includegraphics[width=\textwidth]{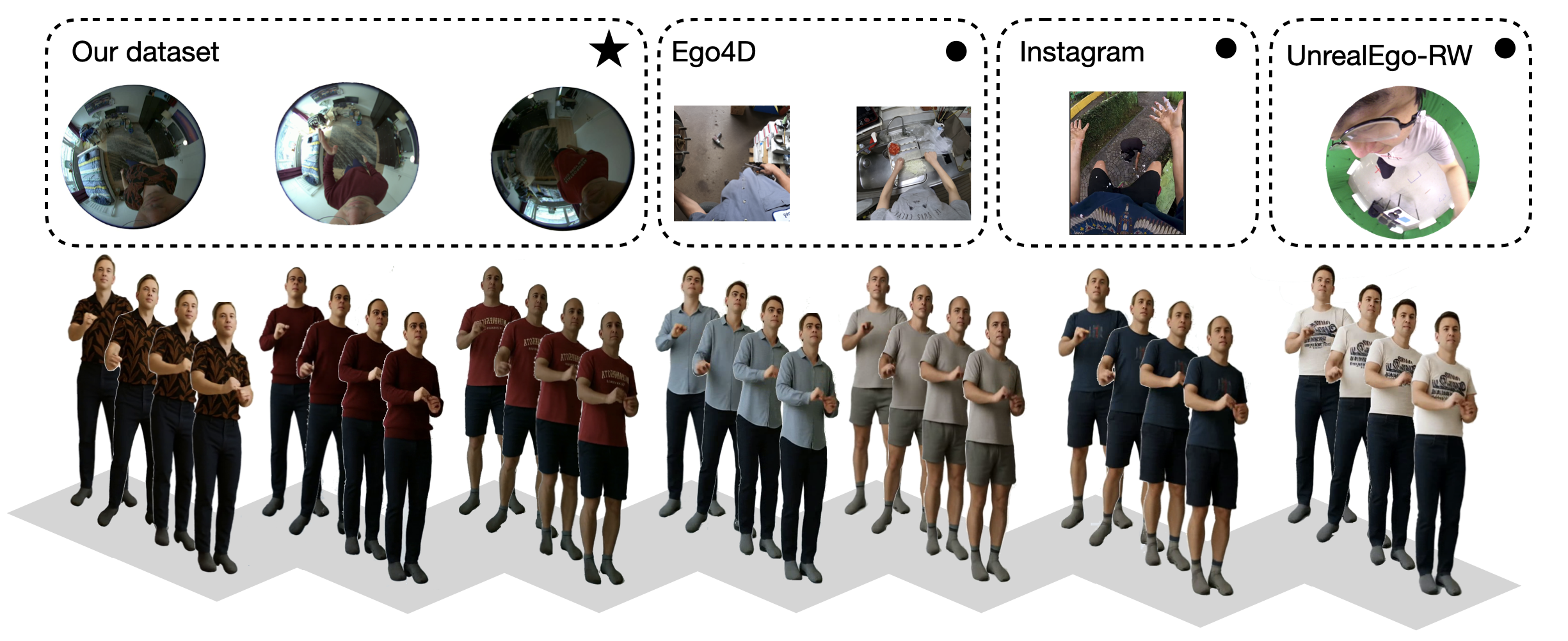}
  \caption{
    \textbf{EgoAnimate} synthesizes a fully animatable avatar from a single egocentric top-down image alone, preserving the subject's clothing while inferring occluded regions such as pants and back views. 
    In the top-row we show unseen egocentric views from our custom dataset, two samples from Ego4D \cite{grauman2022ego4d}, and one sample each from Instagram and from the UnrealEgo2 \cite{hakada2022unrealego, hakada2024unrealego2} dataset.
    Notably, our model has only over seen egocentric views of our dataset ($\star$) during training, meaning that all other input views are out-of-distribution ($\bullet$).
    Nevertheless, our model is capable of producing animatable avatars even from those egocentric images. While one subject appears visually distinct from the training distribution, only body reconstruction is performed and facial regions are not modeled, making this variation negligible for our purposes.
  }
  \label{fig:teaser}
\end{teaserfigure}


\maketitle

\section{Introduction}

VR telepresence is an emerging technology that uses a head-mounted device (HMD) to enable people to communicate and interact with others while experiencing a sense of presence. Recent studies~\cite{rzeszewski2020virtual,barreda2022psychological} indicate the positive effects of this more immersive experience during times of isolation, such as during the Covid-19 pandemic.
To facilitate this feeling of presence, animatable full-body human avatars are an essential component of VR telepresence.
Tools such as VRChat or Meta Quest let users select from a variety of cartoon-style characters while more advanced technologies such as Apple Vision Pro provide scanning technology to create their head \textit{Personas}, which more closely resemble the user, and thus increase the immersiveness of the experience.
To articulate the avatar, controllers or head-mounted cameras (HMCs) integrated into the HMD are used.
Recently, EgoAvatar~\cite{chen2024egoavatar} introduced HMC view-driven full-body avatars with an impressive level of detail, where the motion of the avatars is driven only by the views of binocular HMC.
However, EgoAvatar requires a complex multi-camera studio setup to extract the 3D avatars and cannot create avatars on the fly, using only the HMCs.

Motivated by the recent successes in image- and video-to-avatar methods, we pose the following question: can a fully animatable avatar be created from just the HMD camera inputs alone? 
Crucially, this would offer a simpler and more accessible solution to VR telepresence immersion that could be easy to integrate all while reducing the dependency on large-scale data capture.


However, while HMCs are cost-effective and readily available in existing HMDs, they pose a significant challenge for image-to-avatar pipelines, due to the domain gap from head-mounted cameras to frontal camera views, typically used for image- or video-to-avatar systems, such as ExAvatar~\cite{moon2024expressive}, AnimateAnyone~\cite{hu2024animate} or DreamPose~\cite{karras2023dreampose}.
The domain gap stems from potentially strong lens distortion, unusual camera perspective, and large degree of body occlusion. 
To address this novel and extremely challenging task, we propose a simple yet effective two-step HMC-to-avatar baseline:
As a first step and core contribution we introduce a Diffusion-based canonization network which takes as input an egocentric image and generates an image of that person from a frontal view in T-pose.
This image is then used in the second step for avatar creation.
Our method pipeline is flexible and allows for various methods for avatar creation.

Our egocentric-to-frontal view model draws inspiration from SiTH~\cite{ho2024sith}, but introduces key modifications. 
Specifically, we enhance the original loss function by combining the noise prediction loss with a perceptual loss in image space. 
This helps the model produce more visually plausible reconstructions. 
The model takes a 512×512 top-down frame as input, encodes it using a frozen Stable Diffusion VAE, and then fuses it with noise-injected latent representations derived from ground truth samples. 
A U-Net with trainable transformer blocks performs denoising, guided by ControlNet's~\cite{zhang2023controlnet} pose input and additional cross-attention signals derived from CLIP embeddings\cite{radford2021learning} of the image.

Furthermore, to train the egocentric-to-frontal view model we collect a paired dataset.
It consists of approximately 3,000 top-down RGB images and 300 manually curated frontal-view images, enhanced using an off-the-shelf image diffusion model~\cite{betker2023improving} to improve lighting and visual fidelity. 
The data is sourced from self-recorded indoor videos captured with a head-mounted camera which is attached to a helmet and an external camera for the frontal view. No depth or multi-view sensors are used. Each frontal image is associated with roughly ten corresponding top-down frames, matched via timestamp and body pose.

In our experiments we explore two main directions:
3D-based avatars which allow for faster rendering and 2D video-based avatars which allow for higher visual quality.

For the 3D avatar case we utilize ExAvatar~\cite{moon2024expressive}, a state-of-the-art gaussian splatting~\cite{kerbl2023gaussian} based method.
ExAvatar can be animated using SMPL~\cite{SMPL:2015} sequences.
As ExAvatar requires a video input we expand the generated frontal image into a 360 degrees video, using MagicMan~\cite{he2024magicman}, which generates multi-view human images from a single human image input.
In our experiments we find that this multi-step neural preprocessing, while necessary, addes small artifacts and noise which makes the final 3D avatar have less visual fidelity.

For the 2D avatar case we experiment with state-of-the-art methods, such as UniAnimate~\cite{wang2025unianimate}, StableAnimator~\cite{tu2024stableanimator}, and MimicMotion~\cite{mimicmotion2024} which can directly animate from an image. In our experiments we find that UniAnimate performs best with our synthesized frontal view image.

To summarize, ours contributions are as follows:
\begin{itemize}
    \item We present the first egocentric image to animatable avatar pipeline. Our work is put together from simple building blocks which allows for easy extension and modification.
    \item As part of this pipeline, we introduce a egocentric-to-frontal view synthesis module, which transforms heavily occluded egocentric images into plausible frontal views that can be used to generate full-body avatars.
    \item We experiment with several image-to-animation modules to determine which one best suits our objectives, and outline potential directions for future research in this area.
    \item Finally, we show that it is possible to reach the same end goal—creating usable, animated avatars—from minimal input and limited hardware, by using a new dataset and our simplified pipeline. This offers a more generalizable and accessible approach to solving the egocentric-view-to-avatar problem.
\end{itemize}

\section{Related Work}
\begin{figure*}[t]
    \centering
    \includegraphics[width=\textwidth]{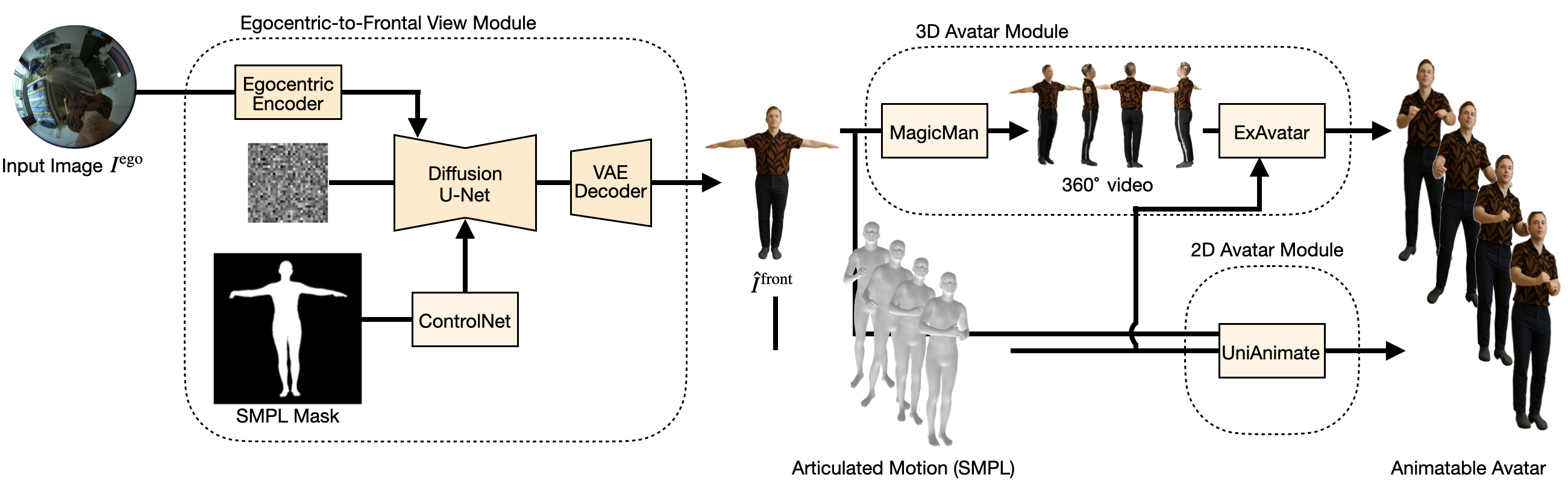}
    \caption{
Method overview of our two-step approach:
Initially, we process an egocentric top-down image $I^{\mathrm{ego}}$ to synthesize a frontal view $\hat{I}^{\mathrm{front}}$ that preserves the subject's clothing. This transformation leverages a pretrained latent diffusion model with ControlNet\cite{zhang2023controlnet} guidance, using the target SMPL\cite{SMPL:2015} mask as conditioning input.
The subsequent phase employs existing techniques to create an animatable avatar from $\hat{I}^{\mathrm{front}}$, controllable via SMPL\cite{SMPL:2015} parameters. Our architecture supports flexible integration with various avatar generation methods, and we evaluate two approaches: the 3D Gaussian Splat-based ExAvatar~\cite{moon2024expressive} and the image-based UniAnimate~\cite{wang2025unianimate}. While UniAnimate directly synthesizes animations from articulated motion and $\hat{I}^{\mathrm{front}}$, ExAvatar requires comprehensive 3D coverage through video input. To address this requirement, we preprocess $\hat{I}^{\mathrm{front}}$ with MagicMan~\cite{he2024magicman} to generate a 360-degree video representation suitable for ExAvatar's avatar creation pipeline.
    }
    \label{fig:methodoverview}
\end{figure*}

\subsection{Diffusion Models}  
Diffusion models have achieved remarkable success in image generation. They have been applied to numerous tasks, including text-to-image generation ~\cite{DBLP:journals/corr/abs-2112-10741, Rombach_2022_CVPR}, image editing ~\cite{brooks2022instructpix2pix, kawar2023imagic, DBLP:conf/iclr/CouaironVSC23}, novel view synthesis \cite{Watson2022NovelVS, Chan_2023_ICCV, Liu_2023_ICCV, kwak2024vivid}, and human body animation \cite{Azadi_2023_ICCV, Xu_2024_CVPR}. Trained on massive amounts of image-text data, they serve as image foundation models and have been used as powerful priors in many domains, particularly when data is scarce \cite{casas2023smplitex, Xiang2023DDM2SD, Song2021SolvingIP}. Notably, Latent Diffusion ~\cite{Rombach_2022_CVPR} enables efficient high-resolution image synthesis by operating in a compressed latent space. Another line of research has focused on improving the controllability of the produced images \cite{BarTal2023MultiDiffusionFD, Li2023GLIGENOG, Mou2023T2IAdapterLA}. ControlNet ~\cite{zhang2023controlnet} and LoRA \cite{Hu2021LoRALA} are very popular methods that have enabled conditioning the generation on a wide variety of modalities, such as depth maps or segmentation maps. In this paper, we base our work on a Stable Diffusion \cite{Rombach_2022_CVPR} model and use ControlNet to control the pose of the generated image. Similarly to SiTH ~\cite{ho2024sith}, we introduce an additional mechanism to condition the denoising process on the input image. 

\subsection{Generative Novel View Synthesis}  
The task of novel view synthesis (NVS) aims at generating images of a scene from unobserved viewpoints, given a set of input images. In this work, we address the task of egocentric-to-frontal view translation. This is a special case of NVS. Significant body occlusions and the use of a single input image make this a particularly challenging setting. Numerous approaches have been proposed to tackle NVS. In the era of deep learning, implicit representations such as Neural Radiance Fields (NeRF) \cite{10.1145/3503250} and, more recently, 3D Gaussian Splatting \cite{kerbl2023gaussian} have achieved remarkable results. However, these methods typically operate in settings where a large number of input images are available and often perform poorly with sparse input. \newline
In the sparse input setting, particularly in the challenging scenario of a single input image, NVS becomes a severely ill-posed problem. In this context, diffusion models have emerged as powerful generative priors, capable of producing plausible generations of unseen parts of the considered objects or scenes. Several works have leveraged diffusion models in various ways. One significant line of work, pioneered by DreamFusion \cite{Poole2022DreamFusionTU}, involves distilling the rich knowledge from pre-trained 2D diffusion models into implicit neural representations like NeRF, using a score distillation loss \cite{Lin2022Magic3DHT, Wang2023ProlificDreamerHA, Tang2023DreamGaussianGG}. Another line of research aims to incorporate 3D priors directly into the diffusion process \cite{Wewer2024latentSplatAV, Lan2024LN3DiffSL}. More recently, enabled by the availability of large-scale synthetic 3D asset datasets such as Objaverse \cite{Deitke2022ObjaverseAU}, several works have adapted 2D diffusion models to directly predict novel views. This approach, pioneered by Zero-1-to-3 \cite{Liu2023Zero1to3ZO}, has shown an impressive level of generalization to unseen objects \cite{Shi2023MVDreamMD, Liu2023One2345AS, Liu2023SyncDreamerGM}.
\newline
Besides general NVS techniques, novel view synthesis for the human body has also received significant attention. Several methods have been developed for this purpose, for example, 3D-aware Generative Adversarial Networks (GANs) \cite{Chan2020piGANPI, 
 Gu2021StyleNeRFAS, Chan2021EfficientG3}, combined with inversion techniques \cite{Xie2022Highfidelity3G, Yuan2023MakeEG, Bhattarai2023TriPlaneNetAE}, have been proposed to generate high-quality novel views of humans from single images. Similar to our method, SiTH \cite{ho2024sith} uses a diffusion model to hallucinate back views from a frontal input. Our specific task of egocentric-to-frontal view translation, however, presents unique challenges causing existing NVS methods, including those designed for humans, to perform poorly. Our paper aims to address this gap.

\subsection{Egocentric Avatars }  
The creation of realistic digital human avatars has been a long-standing and significant challenge in computer vision and graphics. High-fidelity approaches often rely on sophisticated capture systems, such as dense multi-camera rigs, to acquire 3D scans or multi-view video, producing photorealistic results \cite{ Ik2023HumanRFHN, Li2024AnimatableGL, Kirschstein2023NeRSembleMR, Cai2022HuMManM4} . However, these setups can expensive and difficult to scale. Consequently, a considerable line of research has focused on more accessible methods for avatar creation, utilizing data from multi-view images \cite{ Hong20243DCH, Zheng2023GPSGaussianGP, Kwon2024GeneralizableHG, Chen2024MVSplatE3}, monocular videos \cite{ Jiang2022InstantAvatarLA, Alldieck2018DetailedHA, Tang2024GAFGA, Liu2024GVARV}, and even single input images \cite{ lu2025gasgenerativeavatarsynthesis, Saito2019PIFuPI, Hong2023LRMLR, Han2023Highfidelity3H}. To represent body shape and pose, some of these methods are template-free, while others utilize 3D Morphable Models (3DMMs) like SMPL \cite{SMPL:2015}as a prior.
\newline
While human avatars from third-person perspectives have received a lot of attention, the specific domain of egocentric avatars has been comparatively less explored. Much of the existing egocentric research has focused on aspects like 3D human pose estimation \cite{ Akada2022UnrealEgoAN, Tom2019xREgoPoseE3, Tom2020SelfPose3E, Akada20233DHP, Millerdurai2024EventEgo3D3H}, or the creation of egocentric head avatars \cite{ Elgharib2019EgoFaceEF, Jourabloo2021RobustEP, Chen2024ShowYF}. To the best of our knowledge, very few works have attempted the challenging task of generating full-body egocentric avatars. Notably, EgoRenderer \cite{Hu2021EgoRendererRH} tackles this by decomposing the rendering into texture synthesis from fisheye inputs, egocentric pose construction, and a final neural image translation to a target view. Similarly, EgoAvatar \cite{chen2024egoavatar} learns a person-specific drivable avatar, represented by 3D Gaussians from multi-view videos. This model is then animated using a personalized egocentric pose detector trained on data from the same subject.
\newline
These methods have achieved remarkable results, however, they require extensive person-specific data for training their personalized models, making them difficult to scale. Our work aims to overcome this limitation by leveraging strong priors in the form of diffusion models to improve generalization to new subjects.

\begin{figure*}[t]
    \centering
    \includegraphics[width=0.8\textwidth]{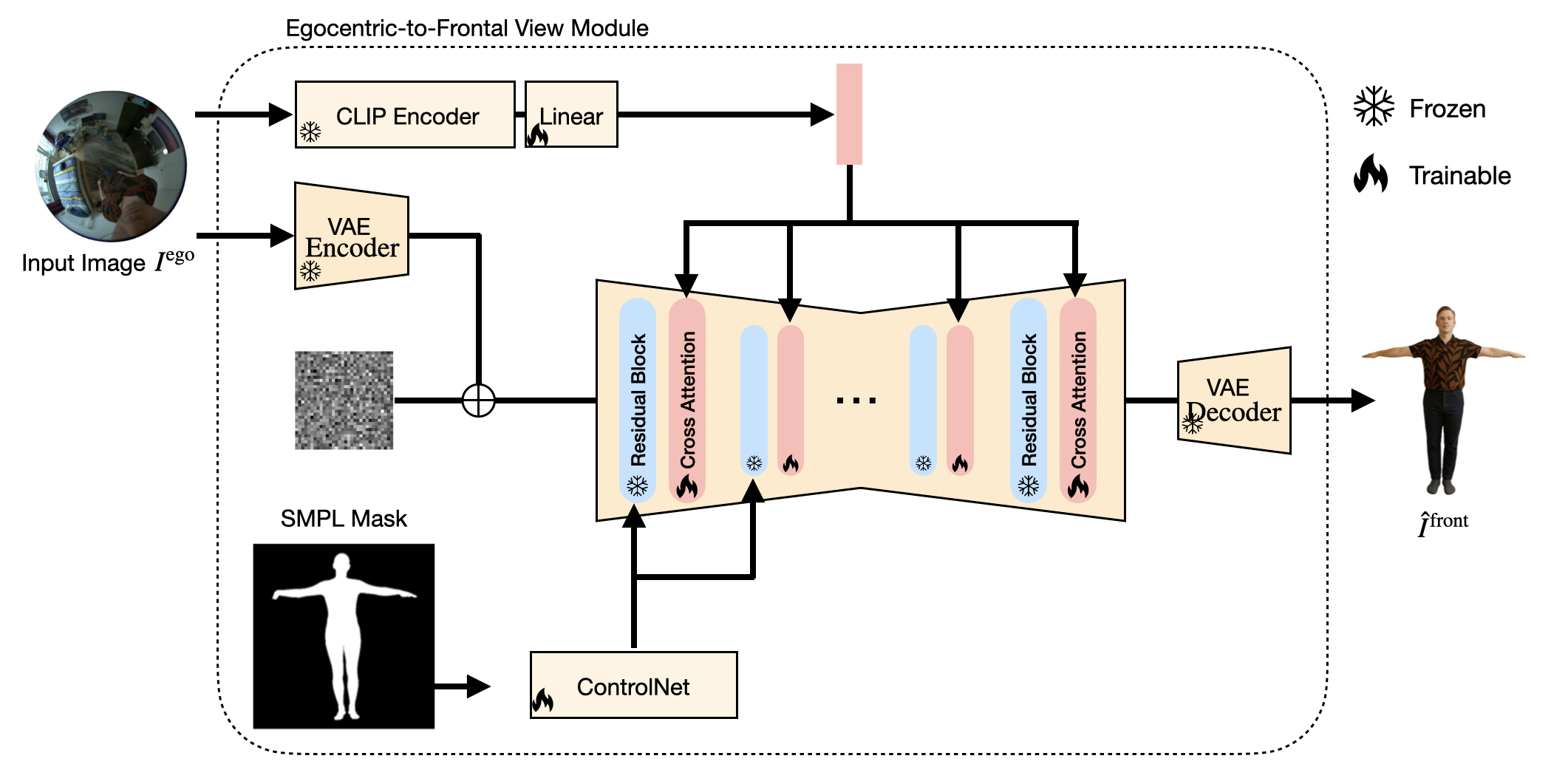}
    \caption{
Overview of our frontal image generation model.
Our model transforms an egocentric top-down image $I^{\mathrm{ego}}$ into a T-pose frontal view $\hat{I}^{\mathrm{front}}$ image through a multi-stream processing pipeline. The input undergoes dual encoding: first via the latent diffusion model's VAE encoder $e_{\mathrm{VAE}}$, where the encoded representation is concatenated with noise signals before the forward diffusion noising process, and second through the CLIP encoder\cite{radford2021learning} $e_{\mathrm{CLIP}}$, which extracts conceptual features subsequently projected linearly to $z^{\mathrm{concept}}$ to accommodate the dimensional requirements of the diffusion model's cross-attention layers.
For structural guidance, we incorporate a target SMPL\cite{SMPL:2015} pose mask into the generation process via ControlNet\cite{zhang2023controlnet}, which provides critical pose information to the diffusion model. The final T-pose frontal image $\hat{I}^{\mathrm{frontal}}$ is synthesized using the VAE decoder $d_{\mathrm{VAE}}$.
In our implementation diagram, \SnowflakeChevron $ $ denotes modules with frozen weights, while \Fire $ $ indicates trainable components.
}
    \label{fig:hmctofrontal}
\end{figure*}

\section{Method}
\label{sec:method}

In this work, we aim to address the challenging task of creating a fully animatable avatar from a single egocentric top-down image.
To do so, we utilize a simple yet effective approach: we finetune a pre-trained latent diffusion model to convert an egocentric image $I^{\mathrm{ego}}$ into a frontal image $\hat{I}^{\mathrm{frontal}}$ which can then be used by off-the-shelf avatar animation methods.
The flexibility of our pipeline allows us to experiment with two avatar paradigms, a 3D geometry approach based on Gaussian splats, and a 2D image-based approach.
An overview of our full avatar creation pipeline can be seen in Figure~\ref{fig:methodoverview}.

In Section~\ref{sec:egotofront} we discuss how to convert an egocentric top-down image to frontal view to bridge the domain gap for avatar creation.
In Section~\ref{sec:avatarcreation} we then discuss how to utilize this generated frontal image for animatable avatar creation.
Finally, we discuss our egocentric-to-frontal image dataset in Section~\ref{sec:dataset}, which is crucial for learning the conversion from egocentric to frontal images.

\subsection{Egocentric-to-Frontal Image Conversion}
\label{sec:egotofront}

Figure~\ref{fig:hmctofrontal} shows a more detailed overview of our Egocentric-to-Frontal image conversion pipeline. Taking the egocentric top-down image as input, it outputs a frontal image in T-pose preserving the clothing of the subject. Facial features are not part of this investigation, as most of the face lies outside the field of view of our camera. The reconstructed front image contains a synthesized face, unrelated to the input subject.

\paragraph{Latent backbone.}
We adopt the latent diffusion framework used in Stable Diffusion ~\cite{Rombach_2022_CVPR}.  
A frozen VAE encoder $\mathcal{E}_{\!\text{VAE}}$ compresses $I_{\!\text{ego}}$ into a latent tensor $z_{\text{ego}}$. The forward diffusion noising process adds Gaussian noise following a linear variance schedule $\beta_t$:
\[
    z_t \;=\; \sqrt{\overline{\alpha}_t}\,z_{\text{ego}} + \sqrt{1-\overline{\alpha}_t}\,\epsilon,
    \quad t\in\{1,\dots,T\},
\]
with $\overline{\alpha}_t=\prod_{s=1}^{t}(1-\beta_s)$ and $\epsilon\!\sim\!\mathcal{N}(0,\mathbf{I})$.

\paragraph{Pose‑conditioned ControlNet.}
A ControlNet branch \cite{zhang2023controlnet} encodes the SMPL\cite{SMPL:2015} pose map into spatial feature maps, which are integrated into the U-Net by adding them directly to the residual streams at both the downsampling and mid-block stages.

\paragraph{CLIP-guided egocentric top-down encoding.}
To extract semantically rich features from the egocentric image $I_{\text{ego}}$, we utilize a pretrained CLIP encoder~\cite{radford2021learning}. The resulting feature vector is projected and spatially expanded to match the resolution of the U-Net’s latent space. These features are then fused with the VAE encoding and passed into the denoising U-Net via cross-attention. This design allows the model to leverage high-level visual semantics from the top-down input, improving its ability to synthesize plausible frontal appearances under severe occlusion.

\paragraph{Objective.}
Egocentric-to-Frontal is supervised by a compound loss:
\[
\mathcal{L}_{\text{Ego2F}} =
\lambda_{\text{diff}} \underbrace{\left\| \epsilon_\theta(\tilde{z}_t, \phi_P, t) - \epsilon \right\|_2^2}_{\mathcal{L}_{\text{diff}}}
+ \lambda_{\text{perc}} \underbrace{\text{LPIPS \cite{zhang2018perceptual}}(\hat{I}_{\text{fr}}, I_{\text{fr}})}_{\mathcal{L}_{\text{perc}}}
\]
with $\lambda_{\text{diff}} = 1$ and $\lambda_{\text{perc}} = 0.2$.

\subsection{Avatar Creation}
\label{sec:avatarcreation}
We considered two different approaches for avatar creation based on our T-pose frontal image - 3D geometry based method and a 2D image based method.

\subsubsection{3D-Geometry-based Avatar}
\label{sec:avatar3d}
The 3D geometry-based avatar creation consists of 2 steps. In the first step we generate a full 360° view of the human subject. To do this, we integrate the pre-trained multi-view synthesis module from MagicMan~\cite{he2024magicman} directly into our pipeline. 
MagicMan is designed to produce consistent multi-view images from a single reference input and its corresponding 3D body mesh, in a single forward pass. We keep all weights frozen and do not fine-tune the model, allowing us to directly benefit from its state-of-the-art multi-view generation capabilities while focusing on improving the egocentric-to-frontal stage of our pipeline.




The second step of the avatar creation results in an animatable 3D avatar by transforming the synthesized multi-view image set. We use the off-the-shelf method ExAvatar~\cite{moon2024expressive}. We do not modify, re-train, or re-implement any part of the ExAvatar pipeline. We directly apply its publicly released codebase on the 20 RGB images from the previous step and their corresponding normal maps.




Our goal in this step was to demonstrate that the frontal view synthesized from an egocentric top-down input enables downstream avatar reconstruction using existing, high-quality tools without any additional training effort. By adopting ExAvatar as a drop-in module, we aim to isolate and evaluate the impact of our contribution: bridging the domain gap between sparse, occluded top-down images and the dense frontal inputs expected by state-of-the-art avatar generation methods. Our results show that once this gap is filled, avatar construction becomes straightforward using existing systems.

\subsubsection{2D-Image-based Avatar}
The 2D image-based avatar creation is a straight-forward application of UniAnimate~\cite{wang2025unianimate} on our frontal T-pose input. UniAnimate synthesizes an animated sequence from the frontal image and some articulated motion.

\subsection{Dataset}
\label{sec:dataset}
We prepared a custom dataset tailored for the Egocentric-to-Frontal generation task. It consists of approximately 3,000 top-down RGB images and 300 manually curated frontal-view images, enhanced using an off-the-shelf image diffusion model \cite{betker2023improving} for realism and semantic fidelity. The data is sourced from self-recorded indoor videos captured with a head-mounted camera which is attached to a helmet and an external camera for the frontal view. No depth or multi-view sensors are used. Each frontal image is associated with roughly ten corresponding top-down frames, matched via timestamp and body pose.

\paragraph{Ground Truth Enhancement.}
Frontal views are sparse and visually enhanced using an off-the-shelf diffusion-based method \cite{betker2023improving} as a post-processing step. While not exact ground-truth, they provide sufficient visual supervision to guide plausible frontal reconstruction, and face, body shape variance in the dataset.

\paragraph{Pre-Processing and Augmentation.}
During training, we apply the following augmentations:

\begin{itemize}
    \item \textbf{Frontal perturbation (with probability $q$):} The frontal image and its corresponding body mask undergo light random transformations, including zoom-in cropping, additional center-shifts, and small rotations. This encourages robustness to variations in framing and pose alignment.
    \item \textbf{Independent top-down rotation (with probability $p$):} To prevent the model from overfitting to absolute positional cues or assuming fixed head/body orientation, we randomly apply a small global rotation to the top-down image independently of its ground-truth counterpart. This ensures that spatial layout is not trivially correlated with the camera-facing direction.
\end{itemize}


\begin{table*}
  \resizebox{\textwidth}{!}{\begin{tabular}{cc|cc|cc|ccc|ccc|ccc}
    \toprule
    \multicolumn{2}{c|}{Model Ablation} &
    \multicolumn{4}{c|}{Egocentric Encoder} &
    \multicolumn{3}{c|}{Full Body} &
    \multicolumn{3}{c|}{Upper Body} &
    \multicolumn{3}{c}{Lower Body} \\
    ControlNet &
    Perc.-Loss &
    SD VAE &
    Sapiens &
    CLIP [CLS] & 
    CLIP grid &
    PSNR$\uparrow$ & SSIM$\uparrow$ & LPIPS$\downarrow$ & 
    PSNR$\uparrow$ & SSIM$\uparrow$ & LPIPS$\downarrow$ & 
    PSNR$\uparrow$ & SSIM$\uparrow$ & LPIPS$\downarrow$ \\
\midrule
& & \cmark & &  & \cmark&
$12.11 \pm 1.22$ & $ 0.7294 \pm 0.03$ & $ 0.2694 \pm 0.03$ &
$12.34 \pm 1.35$ & $0.7949 \pm 0.025$ & $0.1853 \pm 0.027$ &
$11.90 \pm 1.05$ & $0.6589 \pm 0.018$ & $ 0.3188 \pm 0.04$ \\ 
\cmark &  & \cmark & &  & \cmark&
$17.05 \pm 0.89$ & $0.869 \pm 0.0005$ & $0.0987 \pm 0.0001$ &
$17.73 \pm 2.09$ & $ 0.8143 \pm 0.0511$ & $0.1381 \pm 0.0630$ &
$16.81 \pm 0.56$ & $0.8937 \pm 0.012$ & $0.0725 \pm 0.017$ \\ 

\cmark &  &  & \cmark &  &\cmark &
$11.81 \pm 0.60$ & $0.7428 \pm 0.006$ & $0.200 \pm 0.012$ &
$12.15 \pm 0.85$ & $0.8081 \pm 0.015$ & $0.1328 \pm 0.020$ &
$11.50 \pm 0.55$ & $0.6772 \pm 0.008$ & $0.2613 \pm 0.018$ \\ 
\cmark & &&   \cmark & \cmark &  &
$15.66 \pm 0.75$ & $0.7730 \pm 0.005$ & $0.1858 \pm 0.01$ &
$16.04 \pm 0.95$ & $0.8448 \pm 0.012$ & $0.1113 \pm 0.011$ &
$15.30 \pm 0.62$ & $0.6998 \pm 0.018$ & $0.3188 \pm 0.035$ \\ 
\cmark & &\cmark&   \cmark & \cmark &  &
$16.78 \pm 0.67$ & $0.8599 \pm 0.003$ & $0.1073 \pm 0.05$ &
$17.16 \pm 1.68$ & $0.8147 \pm 0.004$ & $0.1456 \pm 0.041$ &
$16.68 \pm 0.62$ & $0.8872 \pm 0.011$ & $0.0928 \pm 0.015$ \\ 
\midrule

\textbf{\cmark} & \cmark & \cmark &  & \cmark &  &
$17.73 \pm 0.97$ & $0.8743 \pm 0.004$ & $0.0835 \pm 0.0006$ &
$18.32\pm 1.97$ & $0.8260 \pm 0.043$ & $0.1174 \pm 0.0429$ &
$17.53 \pm 0.73$ & $0.9026 \pm 0.012$ & $0.0725 \pm 0.017$ \\ 
\bottomrule
  \end{tabular}}
\caption{
Ablation results for different components of our pipeline. Our full model (bottom row) achieves the best performance across all metrics and body regions. We analyze the impact of ControlNet \cite{zhang2023controlnet} conditioning, perceptual loss, Stable Diffusion VAE (SD VAE) \cite{Rombach_2022_CVPR}, custom CLS-based CLIP encoding\cite{radford2021learning}, spatial grid-based CLIP encoding and Sapiens features. Results are reported for full body, upper body, and lower body regions using PSNR, SSIM\cite{wang2004image}, and LPIPS\cite{zhang2018perceptual} metrics.
}
\label{tab:ablation}
\end{table*}

\begin{figure*}[t]
  \centering
  \includegraphics[width=0.65\textwidth]{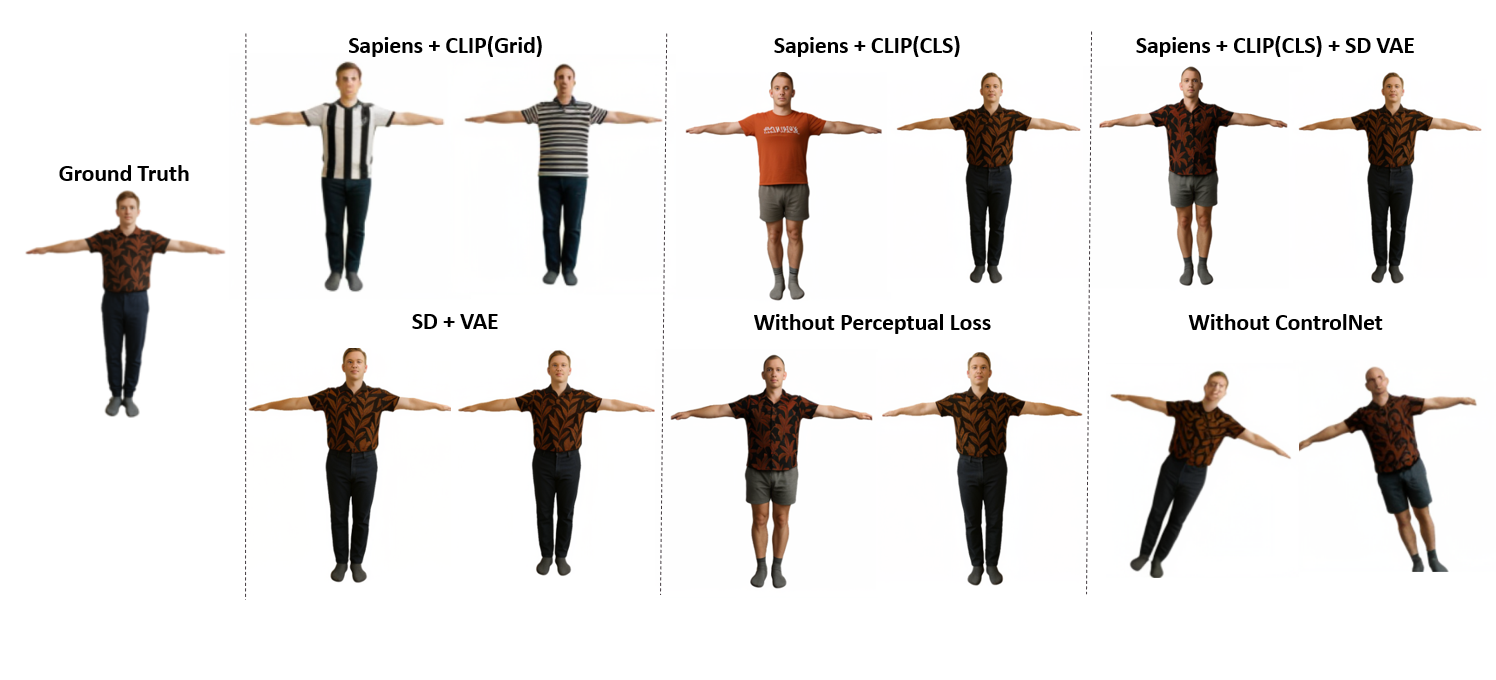} 
  \caption{
    \textbf{Ablations on different Egocentric-to-frontal model configurations.}
    Our method consistently preserves both clothing type and pose across variations, demonstrating robustness compared to alternative configurations. Please zoom in for enhanced view.
  }
  \label{fig:ablations}
\end{figure*}

\section{Experiments}

In this section we evaluate the effects of our model architecture.
In Section \ref{sec:ego2fronteval} we discuss the quantitative evaluation of our egocentric-to-frontal module while Section \ref{sec:avatarcreation} compares various avatar creation methods utility in the context of our method.
Last but not least we discuss qualitative results in Section \ref{sec:qualitative}.


\begin{table}
\centering
\begin{tabular}{cc|cc}
\hline
Perc.-Loss & Encoder & Lower Body (\%) & Upper Body (\%) \\
\hline
 & SD VAE & 62\% & 82\% \\
 & Sapiens & 21\% & 72\% \\
 \hline
\cmark & SD VAE & \textbf{79\%} & \textbf{87\%} \\ 
\hline
\end{tabular}
\caption[]{\textbf{Clothing reconstruction accuracy on unseen samples for top-down to frontal synthesis.} \normalfont{} Percentages reflect how often the generated outfit matches the ground truth in terms of coarse clothing type, such as correctly reproducing shorts versus pants, across 100 samples.}
\vspace{-1.5em} 

\label{tab:percentages_comparison}
\end{table}

\begin{table}[t]
\centering
\small
\setlength{\abovecaptionskip}{3pt}
\setlength{\belowcaptionskip}{3pt}
\begin{tabular}{lccc}
\hline
Option & BS ($\uparrow$) & MR ($\downarrow$) & Runtime ($\downarrow$) \\
\hline
ExAvatar~\cite{moon2024expressive} & 33 & 3.20 & \textbf{1.0} \\
MimicMotion~\cite{mimicmotion2024} & 10 & 3.76 & 2.0 \\
StableAnimator~\cite{tu2024stableanimator} & 86 & 1.90 & 3.75 \\
\textbf{UniAnimate}~\cite{wang2025unianimate} & \textbf{117} & \textbf{1.15} & 22.5 \\
\hline
\end{tabular}
\caption[]{
User study results. Runtime is relative to the fastest method. BS stands for Borda Score and MR stands for Mean Rank.
}
\vspace{-1.2em}
\label{tab:user_study_borda}
\end{table}

\subsection{Egocentric-to-Frontal Module Evaluation}
\label{sec:ego2fronteval}

\noindent \textbf{Dataset}:
We trained our Egocentric-to-Frontal module of EgoAnimate on 540 frontal images collected from our custom dataset. 
The dataset was captured using both head-mounted devices and external cameras to provide paired top-down and frontal views. 
As the egocentric view and frontal image do not have to match in pose we obtain 10 corresponding top-down pose variations to introduce motion diversity and enhance robustness. 
To evaluate the performance of our Egocentric-to-Frontal module, we used 60 samples from the same capture setup. 

\noindent \textbf{Baselines}:
We introduce the novel task of converting egocentric views to frontal T-posed images, establishing the first baseline for this challenging problem. To thoroughly investigate egocentric encoding strategies, we evaluate two alternative approaches:
First, we replace the pre-trained stable diffusion VAE with Sapiens \cite{khirodkar2024sapiens} features—a model with strong human-specific priors—and apply convolutional downsampling to match the required latent resolution. This modification aims to leverage Sapiens' specialized human representation capabilities.
Second, we enhance the CLIP embedding\cite{radford2021learning} pathway by incorporating a learnable transformer decoder that processes the CLIP ViT feature grid. This architecture theoretically enables more focused attention on human elements while suppressing background information.

Contrary to expectations, our experimental results indicate that neither modification yields performance improvements over our original architecture. We analyze the underlying factors for these outcomes in the following sections.

\noindent\textbf{Metrics}:
For our evaluation methodology, we adopt established metrics from the avatar generation literature~\cite{ho2024sith,moon2024expressive,wang2025unianimate} to assess image fidelity. Specifically, we quantify the correspondence between generated and ground-truth frontal images using PSNR (Peak Signal-to-Noise Ratio) for pixel-level accuracy, SSIM\cite{wang2004image} (Structural Similarity Index) for perceptual structural alignment, and LPIPS\cite{zhang2018perceptual} (Learned Perceptual Image Patch Similarity) for feature-space similarity that better correlates with human visual perception.

We further introduce a novel evaluation metric specifically designed for egocentric-to-frontal view generation: coarse clothing type accuracy for both lower and upper body garments. This metric evaluates the model's ability to correctly infer fundamental clothing categories despite the challenging egocentric perspective. \textit{Lower Body \%} measures the accuracy in distinguishing between shorts and pants, while \textit{Upper Body \%} quantifies the accuracy in differentiating between t-shirts and sweaters. These metrics provide crucial insights into the model's capability to preserve semantic clothing attributes when reconstructing frontal views from partial egocentric observations.

\noindent \textbf{Quantitative Results:}
We report our results on PSNR, SSIM\cite{wang2004image} and LPIPS\cite{zhang2018perceptual} in Table \ref{tab:ablation} and our results on clothing reconstitution accuracy in Table \ref{tab:percentages_comparison}.
We find that both ControlNet and Perceptual loss improve overall generation performance, as can be furher seen in Figure \ref{fig:ablations}.

Surprisingly, the strong human prior Sapiens~\cite{khirodkar2024sapiens} is outperformed by the SD VAE.
We hypothesis that this is due to the SD VAE being finetuned for the latent-diffusion architecture while Sapiens does not adhere to those expected distributions.
Furthermore, Sapiens is not trained on egocentric views and thus tend to ignore lower-body parts, such as pants.
This is further confirmed in its low lower-body accuracy in Table \ref{tab:percentages_comparison}.

Similarly, learning features from the CLIP ViT grid\cite{radford2021learning} rather than directly utilizing the CLIP embedding\cite{radford2021learning} did not provide tangible improvements.
We argue that this is due to the model only relying on high level contextual information in the CLIP\cite{radford2021learning} part of the model, for which the CLIP embedding is sufficient.


\subsection{Avatar Animation Module Evaluation}
\label{sec:avatarevaluation}
To animate our reconstructed frontal images, we evaluated four state-of-the-art methods: MagicMan \cite{he2024magicman} + ExAvatar \cite{moon2024expressive} (a pipeline that first generates 360-degree views before creating a 3D avatar), MimicMotion \cite{mimicmotion2024}, StableAnimator \cite{tu2024stableanimator}, and UniAnimate-DiT \cite{wang2025unianimate}. This comparative analysis aimed to identify the most effective approach for generating high-quality animations from egocentric perspectives.

Our evaluation incorporated both computational performance metrics and perceptual quality assessments through a comprehensive user study with 41 participants. Given the absence of ground-truth animations with matching clothing, we employed a ranking-based evaluation protocol where participants ranked animations across five clothing variations according to three criteria: clothing consistency, motion realism, and animation smoothness. To maintain evaluation focus, participants were explicitly instructed to disregard heads and backgrounds while concentrating on motion integrity, limb movement coherence, and overall visual quality.
Interestingly, the overall model performance negatively correlates with the method runtime.

As presented in Table \ref{tab:user_study_borda}, UniAnimate \cite{wang2025unianimate} emerged as the preferred method, followed by StableAnimator \cite{tu2024stableanimator}, ExAvatar \cite{moon2024expressive}, and MimicMotion \cite{mimicmotion2024}. Based on these findings, we implemented UniAnimate \cite{wang2025unianimate} as our primary animation framework for the results presented throughout this paper.

\subsection{Qualitative Results}:
\label{sec:qualitative}
\begin{figure}[t]
  \centering
  \includegraphics[width=1\columnwidth]{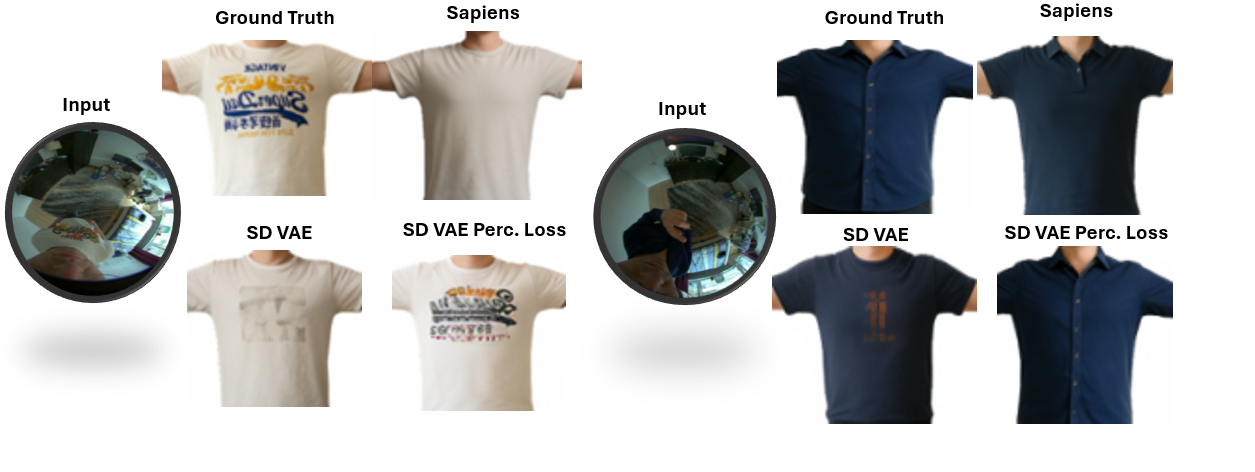}
  \caption{\textbf{Reconstructed Top-Wear:}  Our method more accurately reconstructs printed designs and predicts clothing types compared to ablations. Please zoom in for details.}
  \label{fig:printouts}
  \vspace{-1.0em}  
\end{figure}

We present qualitative results in Figure \ref{fig:teaser}, demonstrating our method's performance on both unseen egocentric views from our dataset and out-of-distribution egocentric images from external sources. These results provide compelling evidence of our approach's generalization capabilities across varied egocentric perspectives.

Notably, our model successfully generates animatable avatars from random samples in the Ego4D~\cite{grauman2022ego4d} dataset despite having no exposure to this data during training. We further evaluate on internet-sourced GoPro footage featuring challenging top-down perspectives of subjects performing dynamic activities like parkour, which presents significant viewpoint and motion complexity. Additionally, we demonstrate effective clothing reconstruction from the UnrealEgo-RW dataset \cite{hakada2022unrealego, hakada2024unrealego2}, further validating our method's robustness across diverse input conditions.

\section{Conclusion}

In this paper, we present a novel modular pipeline that addresses the TopDown-to-Avatar problem using a combination of generative novel view synthesis and off-the-shelf avatar construction tools. By reframing the challenging task of egocentric avatar reconstruction into two simpler tasks- egocentric-to-frontal image translation and frontal image animation-we enable high-quality and efficient full-body avatar generation from a single egocentric image.

Our primary contribution is the development of a TopDown-to-Frontal (T2F) synthesis module built upon Stable Diffusion and ControlNet. This component produces realistic frontal views from top-down inputs, while being robust to both pose variation and occlusion.
We also construct a small, easy-to-use, yet carefully curated dataset of top-down images paired with DALL·E-enhanced frontal views. We further show that by combining our frontal outputs with pre-trained multi-view generation (MagicMan) and avatar modeling (ExAvatar), the full avatar pipeline generalizes to new identities without the need for additional fine-tuning or retraining.

\section{Limitations and Future Work}

While our method demonstrates strong generalization to out-of-distribution egocentric views and clothing styles, it has several limitations that suggest directions for future improvement.

\begin{itemize}
    \item \textbf{Limited Demographic Diversity.} Our dataset primarily includes individuals with similar body types and clothing styles. While the method focuses on reconstructing body shape and pose rather than facial detail, broader demographic representation---including different body shapes, skin tones, and clothing---would improve fairness and robustness.

    \item \textbf{No Face Modeling.} We intentionally exclude facial regions due to occlusion in top-down views. While this simplifies the task and avoids identity leakage, it limits realism in use cases like full telepresence.

    \item \textbf{Dependency on Clothing Visibility.} Although our pipeline infers occluded regions, highly ambiguous clothing geometries (e.g., long coats or flowing garments) can degrade reconstruction quality. Incorporating temporal cues from short top-down video clips may improve consistency.

    \item \textbf{Synthetic Appearance in Certain Cases.} Despite using perceptual and pose losses, the synthesized frontal images can exhibit subtle artifacts, especially in cases with extreme occlusion or unusual body posture. These artifacts are often smoothed out in downstream animation, but improving image realism remains an active research direction.
\end{itemize}


\bibliographystyle{ACM-Reference-Format}


\appendix

\end{document}